\documentclass[journal]{IEEEtran}
\ifCLASSINFOpdf
\else
\fi
\pdfoutput=1
\usepackage{cite}
\usepackage{setspace}
\usepackage{amsmath,amssymb,amsfonts}
\usepackage{algorithmic}
\usepackage{graphicx}
\usepackage{textcomp}
\usepackage{xcolor}
\usepackage{CJKutf8}
\usepackage{color}
\usepackage{amsmath}
\usepackage{amssymb}
\begin{document}
\title{Lightweight Machine Learning for Digital Cross-Link Interference Cancellation with RF Chain Characteristics in Flexible Duplex MIMO Systems}

\author{Jing-Sheng~Tan, Shaoshi~Yang,~\IEEEmembership{Senior~Member,~IEEE}, Kuo Meng, Jianhua~Zhang,~\IEEEmembership{Senior~Member,~IEEE},~Yurong~Tang, Yan~Bu, Guizhen~Wang
\thanks{J.-S. Tan, S. Yang (\textit{Corresponding author}), K. Meng and J. Zhang are with the School of Information and Communication Engineering, Beijing University of Posts and Telecommunications. S. Yang is also with the Key Laboratory of Universal Wireless Communications, Ministry of Education, and J. Zhang is also with the State Key Laboratory of Network and Switching Technology, Beijing 100876, China (E-mail: \{jingsheng.tan, shaoshi.yang, meng\_kuo, jhzhang\}@bupt.edu.cn). Y. Tang, Y. Bu and G. Wang are with the China Mobile
Research Institute, Beijing 100053, China (E-mail: \{tangyurong, buyan, wangguizhen\}@chinamobile.com).}
}

\maketitle
\begin{abstract}
The flexible duplex (FD) technique, including dynamic time-division duplex (D-TDD) and dynamic frequency-division duplex (D-FDD), is regarded as a promising solution to achieving a more \textcolor{black}{flexible} uplink/downlink transmission in \textcolor{black}{5G-Advanced} or 6G mobile communication systems. However, it may introduce serious cross-link interference (CLI). For better mitigating the impact of CLI, we first present a more realistic base station (BS)-to-BS channel model incorporating the radio frequency (RF) chain characteristics, which \textcolor{black}{exhibit} a hardware-dependent nonlinear property, \textcolor{black}{and hence} the accuracy of conventional channel modelling is inadequate for CLI cancellation. \textcolor{black}{Then, we propose a channel parameter estimation based polynomial CLI canceller and} two machine learning (ML) based CLI cancellers \textcolor{black}{that use} the lightweight feedforward neural network (FNN). Our simulation results and analysis show that the ML based CLI cancellers \textcolor{black}{achieve} notable performance improvement and dramatic reduction of computational complexity, \textcolor{black}{in comparison with the polynomial} CLI canceller. 
\end{abstract}

\begin{IEEEkeywords}
Flexible duplex, dynamic TDD/FDD, cross-link interference, 6G, machine learning, MIMO, RF chain.
\end{IEEEkeywords}
\vspace{-5pt}
\section{Introduction}
\textcolor{black}{The 5G-Advanced and 6G} mobile communication systems still have a strong demand for constantly improving the data transmission rate and spectral efficiency, not only on the downlink (DL), which is conventional, but more importantly also on the uplink (UL). This trend is consolidated by the development of UL-centric broadband communication (UCBC), which is an essential paradigm shift for enabling the vertical-industry-oriented wideband Internet of Things (WB-IoT) and is \textcolor{black}{actively} advocated by the 5G-Advanced \textcolor{black}{standardization, as part of the 3GPP Release 18 planned for 2024}. The flexible duplex (FD) technique, including the dynamic time-division duplex (D-TDD) and dynamic frequency-division duplex (D-FDD), is regarded as a promising solution to achieving a more \textcolor{black}{flexible} UL/DL transmission, since it allows the UL/DL transmission direction to be changed dynamically for adapting to the instantaneous traffic variation. 

More specifically, FD allows each cell to have different subframe configurations based on its own service conditions without synchronization, so as to achieve service adaptation. FD is also capable of reducing the outage latency caused by the mismatch between the expected \textcolor{black}{DL-to-UL} traffic ratio and the actual \textcolor{black}{DL-to-UL} traffic ratio. However, there is still a serious challenge that may prevent the realization of potential benefits of FD systems, namely the cross-link interference (CLI).

\textcolor{black}{Since the introduction of} the enhanced interference mitigation and traffic adaptation \textcolor{black}{features into} 3GPP Release 12 (LTE-Advanced), a variety of approaches have been proposed regarding CLI mitigation, e.g., 
\textcolor{black}{cell} clustering schemes \cite{IEEEexample:CLI_Clu}, 
scheduling and resource allocation schemes \cite{IEEEexample:CLI_Reso}, 
power control schemes \cite{IEEEexample:CLI_Pow}, 
and beamforming schemes \cite{IEEEexample:CLI_Beam}. 
In addition, there are schemes based on adjustment of UL and DL configurations \cite{IEEEexample:CLI_UDcong} and joint optimization schemes based on the integration of multiple above-mentioned techniques \cite{IEEEexample:CLI_Joint}. However, these schemes can only decrease the probability of causing CLI, but cannot enable the interfered \textcolor{black}{base station (BS)} to eliminate the impact of CLI when it has occurred. In contrast, the interference cancellation (IC) technique aims to cancel the CLI or reduce its \textcolor{black}{negative impact} and is fundamentally different from the above schemes. The existing CLI cancellation methods depend on channel state information (CSI) \cite{IEEEexample:CLI_UDcong, IEEEexample:CLI_CallbackLink, IEEEexample:R1_1703110_of_3GPP}, whose accuracy, however, has \textcolor{black}{not been given adequate attention} in reconstructing the channel, especially when the modelling of the \textcolor{black}{BS}-to-BS equivalent channel is concerned.

Against the above background, \textcolor{black}{in this paper we first} propose a more realistic BS-to-BS multi-input multi-output (MIMO) channel model incorporating the \textcolor{black}{hardware-dependent nonlinearity characteristics of radio frequency (RF) chain. Then, based on this channel model, a polynomial channel parameter estimation aided digital CLI canceller and} two machine learning (ML) based CLI cancellers \textcolor{black}{that use} the lightweight feedforward neural network (FNN) \textcolor{black}{are proposed}. Our simulation results and analysis demonstrate that the ML based CLI cancellers are more advantageous \textcolor{black}{than the polynomial CLI canceller} in terms of the computational complexity and the CLI cancellation performance.

\vspace{-5pt}
\section{System Model and Problem Formulation}\label{Section2}
\subsection{System Model}
We consider a multi-cell MIMO system operating in the FD (e.g., the D-TDD) mode, as depicted in Fig. 1. The interfered BS (\textcolor{black}{denoted as} BS$_0$) serves $K$ \textcolor{black}{user equipments (UEs)}, and the sets of neighboring 
\begin{figure}[t]
\vspace{-15pt}
\setlength{\abovecaptionskip}{-0.1cm}   
\setlength{\belowcaptionskip}{-0cm}  
\centerline{\includegraphics[width = .38\textwidth]{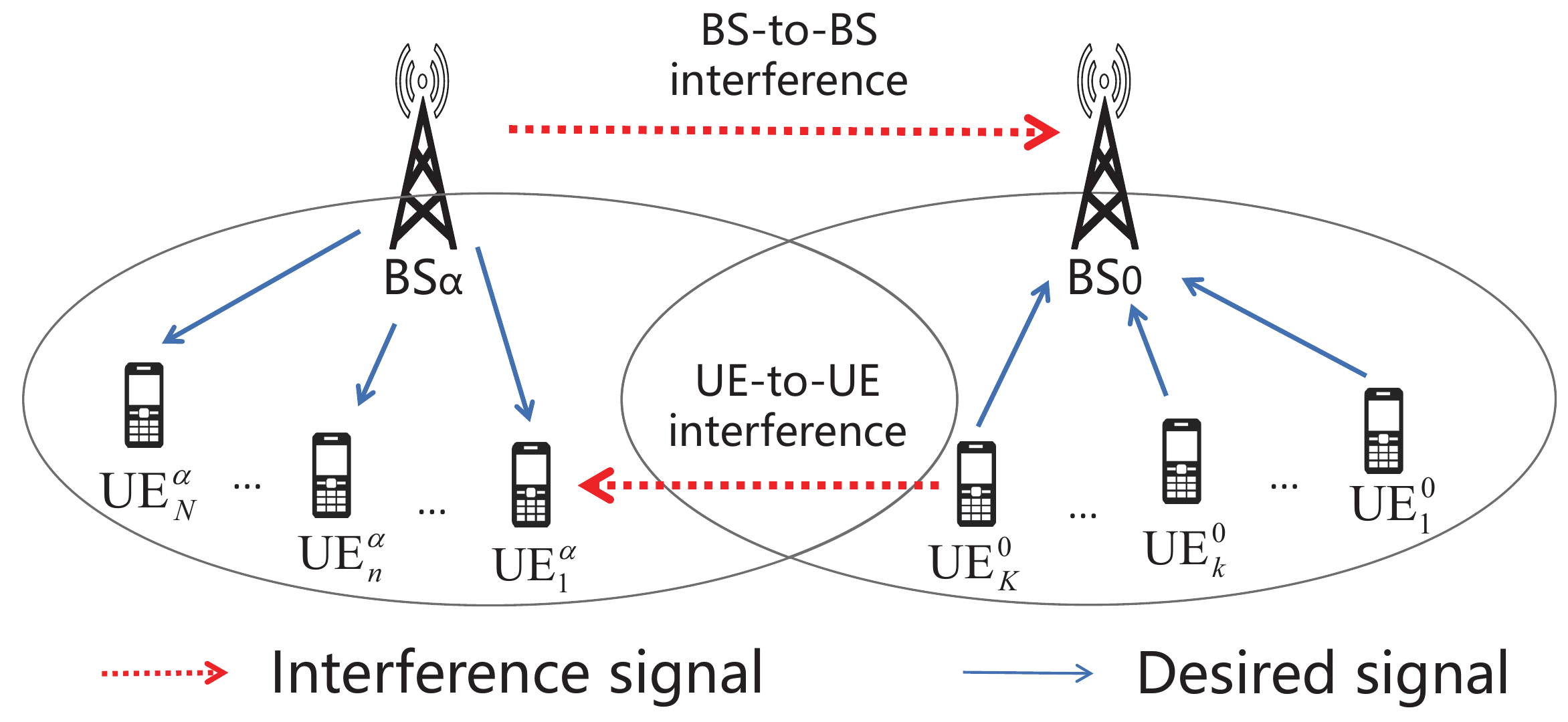}}
\caption{Illustration of CLI in \textcolor{black}{a multi-cell D-TDD MIMO} system.}
\label{B2B}
\vspace{-20pt}
\end{figure}
BSs operating in the UL and DL states are $\mathbb{B}_\text{U}$ and $\mathbb{B}_\text{D}$, respectively. There are two types of CLI. On the UL, the BS$_0$ receives interference from the neighboring BS$_\alpha \in \mathbb{B}_\text{D}$ that is performing DL transmission, and this type of interference is called DL-to-UL interference or BS-to-BS interference. On the DL, in addition to receiving the signal from BS$_\alpha$, UE$^{\alpha}_1$ also receives the UL signal of UE$^0_K$ as interference, which is called UL-to-DL interference or UE-to-UE interference. Due to the low transmission power of UE and the high path-loss between UEs, the UL-to-DL interference is usually much weaker than the DL-to-UL interference. Therefore, in this paper, we focus on the DL-to-UL interference cancellation at the BS side. Suppose the \textcolor{black}{numbers} of antenna elements of UE$^0_k$ and BS$_0$ \textcolor{black}{are} $M_{0,k}$ and $N_{0}$, respectively. The \textcolor{black}{signal ${r}_{0,k}^{n_0}[n]$ received} on the \textcolor{black}{$n_0$th} antenna element of BS$_0$ from UE$^0_k$ at the $n$th sampling instant is expressed as
\begin{equation}\label{eq:signal_model}
\setlength{\abovedisplayskip}{3pt}
\setlength{\belowdisplayskip}{3pt}
{r}_{0,k}^{n_0}[n]=s^{n_0}_{0,k}[n]
\!+\!{s}_\text{D}^{n_0}[n]\!+\!{s}_\text{U}^{n_0}[n]\!+\!{n}_{0,k}^{n_0}[n],
\end{equation}
where $s^{n_0}_{0,k}[n]$ is the UL received signal associated with UE$^0_k$ that is served by BS$_0$, ${s}_\text{D}^{n_0}[n]$ is the overall DL-to-UL interference caused by the BSs in $\mathbb{B}_\text{D}$ (e.g. BS$_\alpha$), and ${s}^{n_0}_\text{U}[n]$ is the overall inter-user interference caused by UEs (e.g. UE$^{\beta}_i$) that are served by the BSs in $ \mathbb{B}_{\text{U}'}$ (e.g. BS$_\beta$). Here \textcolor{black}{we have} $\mathbb{B}_{\text{U}'}=\mathbb{B}_\text{U} \cup$\{BS$_0$\}. In addition, ${n}^{n_0}_{0,k}[n]$ is the additive white Gaussian noise (AWGN).
We model the link between any pair of transmit-receive antennas as a linear finite impulse response (FIR) dispersive channel, and $s^{n_0}_{0,k}[n]$ is expressed as 
\begin{equation}
\setlength{\abovedisplayskip}{2pt}
\setlength{\belowdisplayskip}{2pt}
s^{n_0}_{0,k}[n]=\sum_{m_{0,k}=1}^{M_{0,k}}\sum_{l=0}^{L-1}h^{n_0}_{m_{0,k}}[l]u_{m_{0,k}}[n\!-\!l],
\end{equation}
where $u_{m_{0,k}}[n]$ is the signal transmitted by the $m_{0,k}$th transmit antenna of UE$^0_k$, $h^{n_0}_{m_{0,k}}[l]$ is the channel impulse response of the $l$th path from the $m_{0,k}$th transmit antenna of UE$^0_k$ to the $n_0$th receive antenna of BS$_0$, \textcolor{black}{and $L$ is the maximum number of multi-path components of all links\cite{IEEEexample:CLI_Fifty}. For a given link, $h^{n_0}_{m_{0,k}}[l]$ is set to $0$ when $l$ exceeds its actual number of paths. We assume $h^{n_0}_{m_{0,k}}[l] \!\!=\!\!r^{-\frac{\gamma }{2}}_{0,k,l} \dot{h}^{n_0}_{m_{0,k}}[l]$,} where $r^{-\frac{\gamma }{2}}_{0,k,l}$ \textcolor{black}{characterizes} the path-loss of the $l$th path \textcolor{black}{from} the $m_{0,k}$th antenna of UE$^0_k$ \textcolor{black}{to} the $n_0$th antenna of BS$_0$, with $r_{0,k,l}$ denoting the distance between them and $\gamma$ representing the path-loss exponent, and $\dot{h}^{n_0}_{m_{0,k}}[l]\!\!\sim\!\! \mathcal{CN}(0,1)$ \textcolor{black}{denotes} the small-scale fading of the channel. Note that the impulse response of the other channels considered in this paper \textcolor{black}{obeys} the same model defined here. Similarly, we have 
\begin{equation}
\setlength{\abovedisplayskip}{2pt}
\setlength{\belowdisplayskip}{2pt}
{s}_\text{D}^{n_0}[n]\!
=\!\! \sum_{\text{BS}_\alpha }  s_{\text{D},\alpha}^{n_0}[n]
=\!\! \sum_{\text{BS}_{\alpha} }\sum_{n_{\alpha}=1}^{N_\alpha }\sum_{l=0}^{L-1}h^{n_0}_{n_\alpha}[l]d_{n_\alpha }[n-l],
\end{equation} 
where ${s}_{\text{D},\alpha }^{n_0}[n]$ represents the DL interference imposed by BS$_\alpha$ on BS$_0$, $h^{n_0}_{n_\alpha}[l]$ is the channel impulse response of the $l$th path from the $n_\alpha$th transmit antenna of BS$_\alpha$ to the $n_0$th receive antenna of BS$_0$, and $d_{n_\alpha }[n]$ denotes the signal sent by the $n_{\alpha}$th antenna of BS$_\alpha$. Here $N_\alpha$ is the number of antenna elements of BS$_\alpha$. The UL interference received at BS$_0$ is given by 
\begin{equation}
\setlength{\abovedisplayskip}{0pt}
\setlength{\belowdisplayskip}{3pt}
{s}_\text{U}^{n_0}[n]
\!\!=\!\!  \sum_{\text{BS}_\beta }\sum_{\text{UE}^\beta_i} \mathbf{s}_{\text{U},\beta,i}^{n_0}[n]
\!\!=\!\!  \sum_{\text{BS}_\beta }\sum_{\text{UE}^\beta_i}
\sum_{m_{\beta,i}}^{M_{\beta,i}}\sum_{l=0}^{L-1}h^{n_0}_{m_{\beta ,i}}[l]u_{m_{\beta,i}}[n-l],
\end{equation}
where $h^{n_0}_{m_{\beta ,i}}[l]$ is the channel impulse response of the $l$th path from the $m_{\beta ,i}$th transmit antenna of UE$^\beta_i$ to the $n_0$th receive antenna of BS$_0$, and $u_{m_{\beta,i}}[n]$ denotes the signal sent by the $m_{\beta,i}$th antenna of UE$^\beta_i$ that uses the same frequency band as UE$^0_k$. Obviously, \textcolor{black}{when we have $\beta = 0$}, ${s}^{n_0}_{\text{U},\beta,i}[n]$ represents intra-cell inter-user interference, and when \textcolor{black}{we have $\beta \neq 0$}, ${s}^{n_0}_{\text{U},\beta,i}[n]$ represents inter-cell inter-user interference. 

The inter-cell/intra-cell inter-user interference received on the UL at BS$_0$ can be avoided by the inter-cell interference coordination technique of LTE/NR and eliminated by \textcolor{black}{multiuser joint detection}, respectively. Even if the inter-cell uplink interference could not be avoided in particular circumstances, it can still be mitigated by the multiuser joint detection based coordinated multi-point (CoMP) UL receiver \cite{IEEEexample:T1}. Hence, ${s}_\text{U}^{n_0}[n]$ is ignored in the proposed IC scheme \textcolor{black}{focusing} on mitigating the BS-to-BS interference, and \eqref{eq:signal_model} is \textcolor{black}{simplified as}
\begin{equation}
\setlength{\abovedisplayskip}{1pt}
\setlength{\belowdisplayskip}{1pt}
{r}_{0,k}^{n_0}[n]\!=\!s^{n_0}_{0,k}[n]
\!+\!{s}_\text{D}^{n_0}[n]\!+\!{n}_{0,k}^{n_0}[n].
\end{equation}

\vspace{-0.35cm}
\subsection{CLI Cancellation Problem Formulation}
Suppose that there is a high-capacity backhaul connection (i.e. X2-like interface) between BS$_0$ and BS$_\alpha$ to enable the exchange of inter-cell information associated with the DL transmission of BS$_\alpha$ \cite{IEEEexample:CLI_CallbackLink, IEEEexample:R1_1703110_of_3GPP}. Then, the DL packet, the modulation and coding, the demodulation reference signals, and so forth, can be known at the BS$_0$. Referring to (5), typically \textcolor{black}{the} CSI-based traditional cancellation (TC) schemes of CLI first obtain the \textcolor{black}{estimate} of the channel impulse \textcolor{black}{response} $h^{n_0}_{n_\alpha}[l]$ \cite{IEEEexample:CLI_UDcong, IEEEexample:R1_1703110_of_3GPP}, without considering the RF chain characteristics, 
then reconstruct the BS-to-BS interference $\hat{s}^{n_0}_{\text{D}, \alpha}$, \textcolor{black}{$\forall \ \textrm{BS}_\alpha \in \mathbb{B}_\text{D}$}, and finally \textcolor{black}{subtract} $\hat{s}^{n_0}_{\text{D}}$ from the received signal ${r}^{n_0}_{0,k}$ to obtain the interference-free signal ${\hat{f}}^{n_0}_{0,k}$, which is composed of the UL received signal associated with UE$^0_k$ and the noise, namely
\begin{equation}
\setlength{\abovedisplayskip}{2pt}
\setlength{\belowdisplayskip}{2pt}
{\hat{f}}_{0,k}^{n_0}
={s}_{0,k}^{n_0}+{\bar{n}}_{0,k}^{n_0}
={r}_{0,k}^{n_0}-\hat{s}^{n_0}_{\text{D}}
={r}_{0,k}^{n_0}-\sum_{\text{BS}_\alpha }  \hat{s}_{\text{D},\alpha}^{n_0},
\end{equation}
where ${\bar{n}}_{0,k}^{n_0}$ consists of ${n}_{0,k}^{n_0}$ and the residual CLI. We note that $\hat{x}$ represents the \textcolor{black}{estimate} of $x$ throughout the paper. For more accurately modelling the DL-to-UL interference, a more realistic BS-to-BS channel model incorporating the RF chain characteristics \textcolor{black}{is described below}. 

\begin{figure}[t]
\vspace{-20pt}
\setlength{\abovecaptionskip}{-0.1cm}   
\setlength{\belowcaptionskip}{-5cm}  
\centerline{\includegraphics[width = .4\textwidth]{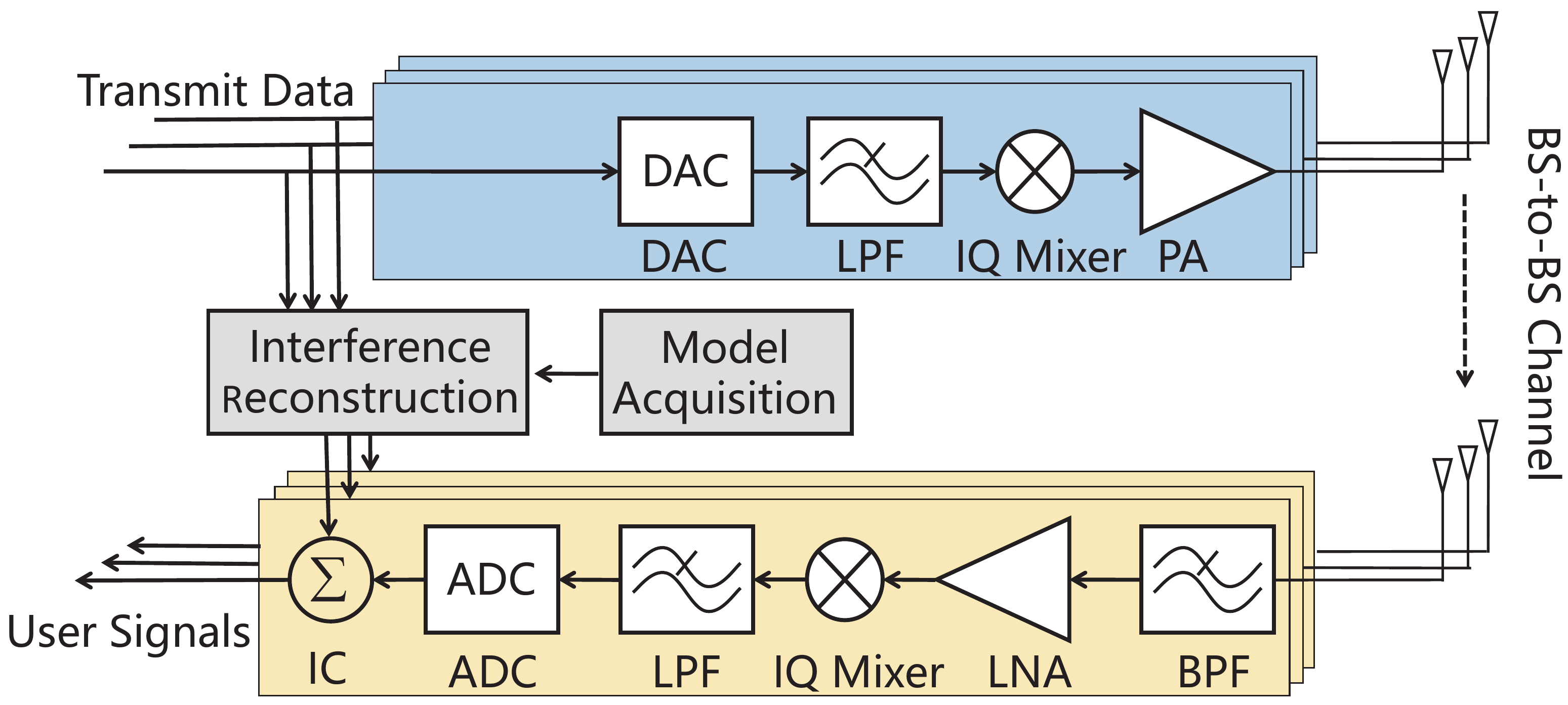}}
\caption{Block diagram of the signal reception and transmission over the BS-to-BS channel. The model acquisition module represents the acquisition of the BS-to-BS channel parameters or the training of the neural networks.}
\label{CLIC}
\vspace{-15pt}
\end{figure}
\vspace{-8pt}
\subsection{BS-to-BS Interference Modelling with RF Chain Characteristics}
Different from the existing contributions on CLI mitigation \cite{IEEEexample:CLI_UDcong, IEEEexample:CLI_CallbackLink, IEEEexample:R1_1703110_of_3GPP}, where \textcolor{black}{the BS-to-BS CSI matrix only} comprises the path-loss and fast fading, in this paper we establish a more realistic \textcolor{black}{BS-to-BS MIMO channel model that considers} the major characteristics of the RF chain to perform the BS-to-BS interference modelling, as shown in Fig. 2. This approach accounts for the effects of both the channel linearity and \textcolor{black}{the} nonlinearity associated with the RF chain. For simplifying the analysis, we consider the impacts of the in-phase and quadrature (IQ) mixer and the power amplifier (PA), while assuming ideal digital-to-analog converter (DAC), low pass filter (LPF), band pass filter (BPF), low noise amplifier (LNA) and analog-to-digital converter (ADC). 

\textcolor{black}{With respect to $d_{n_\alpha}[n]$ in Eq.~(3) and the MIMO transceiver illustrated in Fig. \ref{CLIC}}, the digital baseband signal passed through the DAC, LPF and IQ mixer is modelled as 
\begin{equation}
\setlength{\abovedisplayskip}{2pt}
\setlength{\belowdisplayskip}{2pt}
d_{n_\alpha}^{\mathrm{IQ}}[n]=K_{1,n_\alpha} d_{n_\alpha}[n]+K_{2,n_\alpha} d_{n_\alpha}^{*}[n],
\end{equation}
where \textcolor{black}{we have} $\left| K_{1,n_\alpha} \right|>>\left| K_{2,n_\alpha} \right|$, $K_{1, n_\alpha}\!=\!1 / 2(1+g_{n_\alpha} e^{j \varphi_{n_\alpha}})$, and $K_{2, n_\alpha}\!=\!1 / 2\left(1-g_{n_\alpha} e^{j \varphi_{n_\alpha}}\right)$, whilst $g_{n_\alpha}$ and $\varphi_{n_\alpha}$ are the gain and phase imbalance parameters \textcolor{black}{of the $n_\alpha$th transmit antenna element, respectively}. \textcolor{black}{The symbol ``*"} denotes complex conjugate. The output signal of the IQ mixer is amplified by the PA, and the response of the PA is approximated \textcolor{black}{by} using the parallel Hammerstein (PH) model \cite{IEEEexample:CLI_PAmodel} \textcolor{black}{and} given as 
\begin{equation}
\setlength{\abovedisplayskip}{2pt}
\setlength{\belowdisplayskip}{2pt}
d_{n_\alpha}^{\mathrm{PA}}[n]\!\!=\!\!\!
\sum_{p=1 \atop p \text { odd }}^{P} \!\!\sum_{m=0}^{M} \!v_{n_\alpha,p}^{\mathrm{PA}}[m] d_{n_\alpha}^{\mathrm{IQ}}[n\!\!-\!\!m] \left| d_{n_\alpha}^{\mathrm{IQ}}[n\!\!-\!\!m] \right|^{p\!-\!1}\!,
\end{equation}
where $M$ is the memory length of the PA, $P$ is the \textcolor{black}{order of nonlinearity considered in the PH} model, and $v_{n_\alpha,p}^{\mathrm{PA}}$ is the impulse response of the $p$th-order nonlinearity. The baseband signal received by the $n_0$th \textcolor{black}{receive} antenna of BS$_0$ at the $n$th sampling instant from \textcolor{black}{the $n_{\alpha}$th antenna of BS$_\alpha$} is given by
\begin{equation}
\setlength{\abovedisplayskip}{2pt}
\setlength{\belowdisplayskip}{2pt}
s^{n_0}_{\text{D},\alpha}[n]=\sum_{n_\alpha=1}^{N_\alpha} \sum_{l=0}^{L-1} h^{n_0}_{n_\alpha}[l] d_{n_\alpha}^{\mathrm{PA}}[n-l].
\end{equation}
By substituting (7) and (8) into (9), we have
\begin{equation}
\setlength{\abovedisplayskip}{1pt}
\setlength{\belowdisplayskip}{1pt}
\begin{split}
s_{\text{D},\alpha}^{n_0} [n]=\sum_{n_\alpha=1}^{N_{\alpha}}\sum_{p=1 \atop p \text { odd }}^{P} & \sum_{q=0}^{p} \sum_{m=0}^{M+L-1} \omega^{n_0}_{n_\alpha,p, q}[m] \\& 
\times \Phi (d_{n_\alpha} [n-m],p,q),
\end{split}
\end{equation}where $\omega^{n_0}_{n_\alpha,p, q}[m]$ \textcolor{black}{characterizes the overall effect of the corresponding coefficients in (7), (8) and (9)}, \textcolor{black}{and $\Phi (x,p,q)$ equals $x^q(x^{*})^{p-q}$}. Therefore, at the $n$th sampling instant, the CLI $\mathbf{s}_{\text{D},\alpha }$ from BS$_{\alpha}$ to BS$_0$ can be expressed as
\begin{equation}
\setlength{\abovedisplayskip}{1pt}
\setlength{\belowdisplayskip}{1pt}
\mathbf{s}_{\text{D},\alpha }[n]=\left [ s^{1}_{\text{D},\alpha}[n],\cdots,  s^{n_0}_{\text{D},\alpha}[n],\cdots, s^{N_0}_{\text{D},\alpha}[n]\right ].
\end{equation}
\vspace{-15pt}
\section{Digital CLI Canceller}\label{Section3}
\vspace{-5pt}
In this section, we propose three types of digital-domain CLI cancellers based on the channel model described by (10), namely the polynomial canceller (PC), the neural network canceller (NNC) and the hybrid canceller (HC). Note that both the NNC and the HC are based on ML. The metric we use to measure the performance of these CLI cancellers is $C_{\mathrm{dB}}$ (expressed in dB), which is defined over the time-window with length $N$ as
\begin{equation}
\setlength{\abovedisplayskip}{3pt}
\setlength{\belowdisplayskip}{3pt}
C_{\mathrm{dB}}\!\!=\!\!10 \textcolor{black}{\log_{10}} \left(\frac{\sum_{n_0=1}^{N_0}\sum_{n=0}^{N-1}|s^{n_0}_{\text{D},\alpha}[n]|^{2}}{\sum_{n_0=1}^{N_0}\sum_{n=0}^{N-1}|s^{n_0}_{\text{D},\alpha}[n]-\hat{s}^{n_0}_{\text{D},\alpha}[n]|^{2}}\right).
\end{equation}
\vspace{-15pt}
\subsection{PC}
When the BS-to-BS MIMO channel with the RF chain characteristics is combined with the interference cancellation philosophy, a CLI canceller that deals with the estimation of the coefficients of a polynomial, as seen in (10),  is conceived. We call it the PC scheme, where the reconstructed interference $\hat{s}^{n_0}_{\text{D},\alpha}$ is obtained by using the data $d_{n_{\alpha}}$ of BS$_{\alpha}$ and estimating the channel parameters $\hat \omega^{n_0}_{n_\alpha,p, q}$. We note that reliable backhaul connections are assumed between BSs, and through BS coordination \textcolor{black}{a ``clean" period exists, during which there is only a single interfering BS and no uplink transmission for the BS$_0$. Hence} BS$_0$ is able to obtain $d_{n_\alpha}[n]$ and $s^{n_0}_{\text{D},\alpha} [n]$ for estimating $\hat \omega^{n_0}_{n_\alpha,p, q}$ with the least squares (LS) method, before the normal data transmission takes place. Note that the NNC \textcolor{black}{and the HC} can \textcolor{black}{also} obtain the training data through \textcolor{black}{this} method.

For a CLI canceller both the number of real-value parameters to be estimated and the computational complexity are important metrics. Floating-point multiplication is faster than floating-point addition, while integer addition is faster than integer multiplication\cite{IEEEexample:IEEE_754_FP}. How much faster depends on the computer architecture, as confirmed by our extensive tests. We simply define the computational complexity as the sum of the real-value addition and multiplication operations required to reconstruct the interference signals. Since the estimation of the parameters \textcolor{black}{$\hat \omega^{n_0}_{n_\alpha,p, q}$} can be done \textcolor{black}{offline} before the normal data transmission takes place, we ignore the computational complexity of estimating \textcolor{black}{$\hat \omega^{n_0}_{n_\alpha,p, q}$}. Since the receiver of BS$_0$ has $N_0$ antenna elements, the number of real-value parameters to be estimated in the PC approach is given as
\begin{equation}
\setlength{\abovedisplayskip}{3pt}
\setlength{\belowdisplayskip}{3pt}
n_{\text{PC}}=0.5N_0N_{\alpha}(M+L)\left({P+1}\right)\left({P+3}\right).
\end{equation}

Addition of two complex numbers requires two real-value additions, and multiplication of two complex numbers requires four real-value multiplications, \textcolor{black}{as well as one addition and one subtraction.} \textcolor{black}{Since modern computers use the same circuits and instructions to handle addition and subtraction, the computational complexities of addition and subtraction are regarded the same.} We ignore the computational complexity of complex conjugation, hence in total we need $6^p$ operations for calculating $\omega^{n_0}_{n_\alpha,p, q}[m] \!\!\times\!\! \Phi (x,p,q)$. As seen from (10) \textcolor{black}{and (11)}, the computational complexity of PC is then expressed as
\begin{equation}
\begin{aligned}
c_\text{PC}\!\!=\!\!N_0N_{\alpha}(M\!\!+\!\!L)(\frac{(35P\!\!+\!\!33)6^{P\!+\!2}\!\!+\!\!12}{35^2}\!+\! \frac{(P\!\!+\!\!1)(P\!\!+\!\!3)}{2} )\!\!-\!\!\textcolor{black}{2N_0}.
\end{aligned}
\end{equation}

A \textcolor{black}{larger} $P$ represents a better fit to the channel nonlinearity, but the computational complexity grows quickly with $P$, which is unbearable on a BS with limited computation resources. We can utilize ML to solve this contradiction, as presented below. Note that since different interference cancellers with different parameters can be trained by the interfered BS$_0$ in the multi-interfering-BS scenario independently and used in parallel, $n_{\text{PC}}$ and $c_{\text{PC}}$ of PC \textcolor{black}{grow} linearly with the number of BS$_\alpha$. This conclusion also holds for the NNC and the HC, and our scheme can remain \textit{lightweight} in \textcolor{black}{more-complex} scenarios.

\vspace{-5pt}
\subsection{NNC}
Feedforward neural network (FNN) is a multi-layer neural network, and \textcolor{black}{it can be trained} by splitting the data into mini-batches and executing a gradient descent update after processing each mini-batch. One pass through the entire training data set is called a training epoch.

The input $\textcolor{black}{\mathbf{x}[n]} \in \mathbb{R}^{2N_{\alpha}(M_{\alpha}+1)}$ of NNC is given by $[\Re{(d_{1}[n])},\Im{(d_{1}[n])},\cdots,\Re{(d_{n_{\alpha }}[n\!-\!m])}, \Im{(d_{n_{\alpha }}[n\!-\!m])},\cdots,\\ \Re{(d_{N_{\alpha }}[n\!-\!M_{\alpha}])}, \Im{(d_{N_{\alpha }}[n\!-\!M_{\alpha}])}]$, and the corresponding ideal output $\mathbf{y}_{\text{NNC}}[n] \in \mathbb{R}^{2N_{0}}$ is expressed as $[\Re{(s^{1}_{\text{D},\alpha }[n])},\\ \Im{(s^{1}_{\text{D}, \alpha }[n])},\cdots,\Re{(s^{N_0}_{\text{D},\alpha }[n])},\Im{(s^{N_0}_{\text{D},\alpha }[n])}]$. To improve the performance and training stability of the FNN, \textcolor{black}{$\mathbf{x}[n]$} is normalized to \textcolor{black}{$\mathbf{x}'=\mathbf{x}[n]' \!\!=\!\! {\mathbf{x}[n]}/{m_1} $} and then fed into FNN, as shown in Fig. 3. Since FNN is trained with the label data \textcolor{black}{$\mathbf{y}_{\text{NNC}}[n]'=\mathbf{y}_{\text{NNC}}[n]/{m_2^{\text{NNC}}}$}, the output $\mathbf{\hat{y}}_{\text{NNC}}[n]'$ of FNN  corresponding to $\mathbf{x}[n]'$ needs to be denormalized to \textcolor{black}{$\mathbf{\hat{y}}_{\text{NNC}}[n]=m_2^{\text{NNC}} \mathbf{\hat{y}}_{\text{NNC}}[n]'$} before interference cancellation. Note that $m_1$ and \textcolor{black}{$m_2^{\text{NNC}}$} are the maximum values of $|d_{n_\alpha}[n]|$ and $|s^{n_0}_{\text{D},\alpha}[n]|$, respectively.

\begin{figure}[t]
\vspace{-10pt}  
\centerline{\includegraphics[width = .42\textwidth]{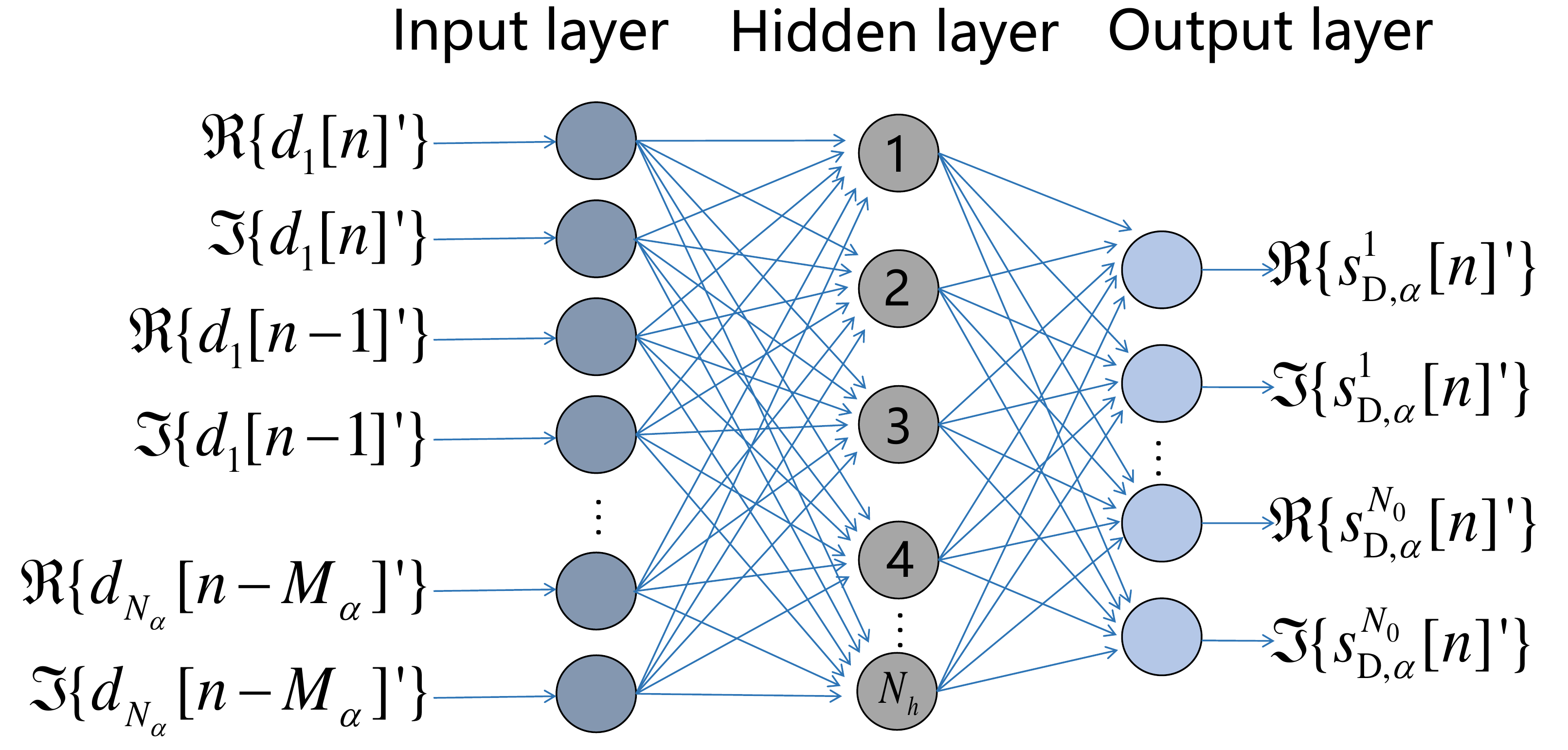}}

\setlength{\abovecaptionskip}{0cm}   
\setlength{\belowcaptionskip}{-3cm}   

\caption{An FNN with $2N_{\alpha}(M_{\alpha}+1)$ input nodes, $N_h$ hidden nodes, and $2N_0$ output nodes, where \textcolor{black}{we have}  $M_{\alpha}=M+L-1$.}
\vspace{-18pt}
\label{NN}
\end{figure}

The parameters of the FNN considered are all real-valued. In the three-layer FNN shown in Fig. 3, we suppose \textcolor{black}{the numbers of nodes on the input layer, hidden layer and output layer are} $N_\text{in}$, $N_\text{h}$ and $N_\text{out}$, respectively. The function from the input layer to the hidden layer can be expressed as $\mathbf x_\text{h}\!\!=\!\!\sigma(\mathbf W_\text{h} \textcolor{black}{\mathbf x'}+\mathbf b_{\text{h}})$, where $\mathbf W_{\text{h}} \in \mathbb{R}^{N_\text{h}\times N_\text{in}}$ is the parameter matrix, $\mathbf b_\text{h} \in \mathbb{R}^{N_\text{h}}$ is the bias, \textcolor{black}{and $\sigma(x)$ represents} the nonlinear activation function, which enables the FNN to well fit the nonlinear relationship between the inputs and the outputs. Therefore, there are \textcolor{black}{totally $(N_\text{in}+1)N_\text{h}$ parameters in} $\mathbf W_\text{h}$ and $\mathbf b_{\text{h}}$. \textcolor{black}{In addition, $N_\text{in}N_\text{h}$ multiplications, $N_\text{in}N_\text{h}$ additions and $c_{\sigma}N_\text{h}$ comparison operations in $\mathbb{R}$ are required for calculating $\sigma(\mathbf W_\text{h} \textcolor{black}{\mathbf x'}+\mathbf b_{\text{h}})$, where $\textcolor{black}{c_{\sigma}}$ represents the complexity of calculating $\sigma(x)$. In this paper, we choose} $\sigma(x)=\max(0,x)$, which is a numeric comparison operation, so we set $\textcolor{black}{c_{\sigma}}=1$. Similarly, without considering the activation function, the number of parameters and the computational complexity of calculation from the hidden layer to the output layer are $(N_\text{h}+1)N_\text{out}$ and $2N_\text{h}N_\text{out}$, respectively. Besides, $N_\text{in}$ and $N_\text{out}$ multiplications, as well as two parameters \textcolor{black}{$1/m_1$} and \textcolor{black}{$m_2^{\text{NNC}}$,} are needed for normalization and denormalization, respectively.

Based on (10) and (11), \textcolor{black}{we have} $N_\text{in}=2N_{\alpha}(M+L)$ and $N_\text{out}=2N_0$. \textcolor{black}{Moreover}, $N_\text{in}$ belongs to hyperparameters that are tunable. Therefore, the number of parameters \textcolor{black}{$n_{\text {NNC}}$} and the computational complexity \textcolor{black}{$c_{\text {NNC}}$} of NNC are given by
\begin{equation}
\setlength{\abovedisplayskip}{3pt}
\setlength{\belowdisplayskip}{1pt}
n_{\text {NNC}}=N_\text{h}[2(M+L)N_\alpha+2N_0+1]+2N_0+2,
\end{equation} 
\begin{equation}
\setlength{\abovedisplayskip}{2pt}
\setlength{\belowdisplayskip}{-2pt}
c_{\text {NNC}}=2(2N_\text{h}+1)[N_\alpha(M+L)+N_0]+\textcolor{black}{c_{\sigma}}N_\text{h},
\vspace{-5pt}
\end{equation}
\textcolor{black}{respectively, while} $n_{\text {NNC}}$ and $c_{\text {NNC}}$ are independent of the nonlinearity order $P$.

\vspace{-10pt}
\subsection{HC}
The CLI signal of (10) can be decomposed as
\begin{equation}
\setlength{\abovedisplayskip}{1pt}
\setlength{\belowdisplayskip}{1pt}
s_{\text{D},\alpha}^{n_0} [n] = s^{n_0,\text{lin}}_{\text{D},\alpha}[n]+s^{n_0,\text{nl}}_{\text{D},\alpha}[n],
\end{equation}
where $s^{n_0,\text{lin}}_{\text{D},\alpha}[n]$ and $s^{n_0,\text{nl}}_{\text{D},\alpha}[n]$ represent the linear part and nonlinear part of $s_{\text{D},\alpha}^{n_0} [n]$, respectively. The linear part corresponds to the terms when \textcolor{black}{we have} $P=1$ and $q=1$ in Eq. (10), and the nonlinear part corresponds to the rest. \textcolor{black}{If} the linear part is eliminated and the nonlinear part is ignored, HC degenerates into TC. \textcolor{black}{The HC approach can be divided into two steps. First, the TC is invoked to estimate the  linear part as $\hat{s}^{n_0,\text{lin}}_{\text{D},\alpha}[n]$. Second, the NNC is used to estimate the remaining nonlinear part as $\hat{s}^{n_0,\text{nl}}_{\text{D},\alpha}[n]\!\!=\!\!s^{n_0}_{\text{D},\alpha}[n]\!-\!\hat{s}^{n_0,\text{lin}}_{\text{D},\alpha}[n]$. In other words, FNN is trained with the input data $\mathbf{x}[n]' \!\!=\!\! {\mathbf{x}[n]}/{m_1}$ and the label data $\mathbf{y}_{\text{HC}}[n]'\!=\! \mathbf{y}_{\text{HC}}[n]/{m_2^{\text{HC}}}$, where $\mathbf{y}_{\text{HC}}[n]$ is expressed as $[\Re{(\hat{s}^{1,\textrm{nl}}_{\text{D},\alpha }[n])}, \Im{(\hat{s}^{1,\textrm{nl}}_{\text{D}, \alpha }[n])},\cdots, \Re{(\hat{s}^{N_0,\textrm{nl}}_{\text{D},\alpha }[n])},\Im{(\hat{s}^{N_0,\textrm{nl}}_{\text{D},\alpha }[n])}]$ and $m_2^{\text{HC}}$ is the maximum value of $|\hat{s}^{n_0,\textrm{nl}}_{\text{D},\alpha}[n]|$.} We note that, because of the low power of the nonlinear part, reconstructing CLI signals \textcolor{black}{by} only using ML may cause the nonlinear part to be ignored.
\textcolor{black}{Similarly,} the number of parameters and the computational complexity of the HC can be expressed respectively as
\begin{equation}
\setlength{\abovedisplayskip}{1pt}
\setlength{\belowdisplayskip}{1pt}
n_{\text {HC}}=2N_0N_{\alpha}(M+L)+n_{\text {NNC}},
\end{equation}
\begin{equation}
c_{\text {HC}}=8N_0N_{\alpha}(M+L)-\textcolor{black}{2N_0}+c_{\text {NNC}}.
\end{equation}

\vspace{-10pt}
\section{Experimental Results}\label{Section4}
In this section, we present the numerical results to compare the performance of the PC, the NNC and the HC methods. 

\begin{table}[b]
\vspace{-15pt}
\footnotesize
\caption{Relevant Parameters of BS-to-BS Transmission}
\vspace{-20pt}
\begin{center}
\begin{tabular}{|c|c|}
\hline
\textbf{Parameters}&{\textbf{Values}} \\
\hline
{Received CLI power}&{-52.1 dBm} \\
\hline
{Transmitting power of the interfering BS}&{47 dBm} \\
\hline
{AWGN power}&{-90 dBm} \\
\hline
{Waveform}&{OFDM} \\
\hline
{Bandwidth}&{13MHz}\\
\hline
{Sampling frequency}&{120MHz}\\
\hline
{ADC resolution}&{12} \\
\hline
{Number of TX/RX antennas}&{4/4} \\ 
\hline
{Fading distribution}&{Rayleigh} \\ 
\hline
{Memory length of the PA ($M$)}&{2} \\
\hline
{Maximum number of multi-path components ($L$)}&{7} \\
\hline
{Nonlinearity order of the interference canceller ($P$)}&{3}\\
\hline
{Batch size of the FNN}&{32}\\
\hline
{Learning rate of the FNN}&{$2\times10^{-4}$}\\
\hline
{Loss function of the FNN}&{Mean squared error}\\
\hline
{Optimization algorithm of the FNN}&{Adam}\\
\hline
\end{tabular}
\label{Simu}
\end{center}
\vspace{-10pt}
\end{table}
\vspace{-10pt}  
\subsection{Data Set}
We study the performance of the PC, the NNC and the HC methods through system level simulations of a D-TDD MIMO system, and the main simulation parameters are shown in Table \ref{Simu}. For the sake of simplicity, the number of BSs \textcolor{black}{is} set to 2. 
The interfered BS obtains 50000 baseband samples $d_{n_\alpha}$ from the interfering BS per transmit antenna via the backhaul connection, and obtains the corresponding CLI $s^{n_0}_{\text{D},\alpha}$ per receive antenna relying on its own ADC. We note that 80\% data are used for training and 20\% data for testing.
\vspace{-10pt}
\subsection{Numerical Results}\label{Sc4c}
\begin{figure}[t]
\vspace{-20pt}
\setlength{\abovecaptionskip}{-0.1cm}   
\setlength{\belowcaptionskip}{-5cm}  
\centerline{\includegraphics[width = .48\textwidth]{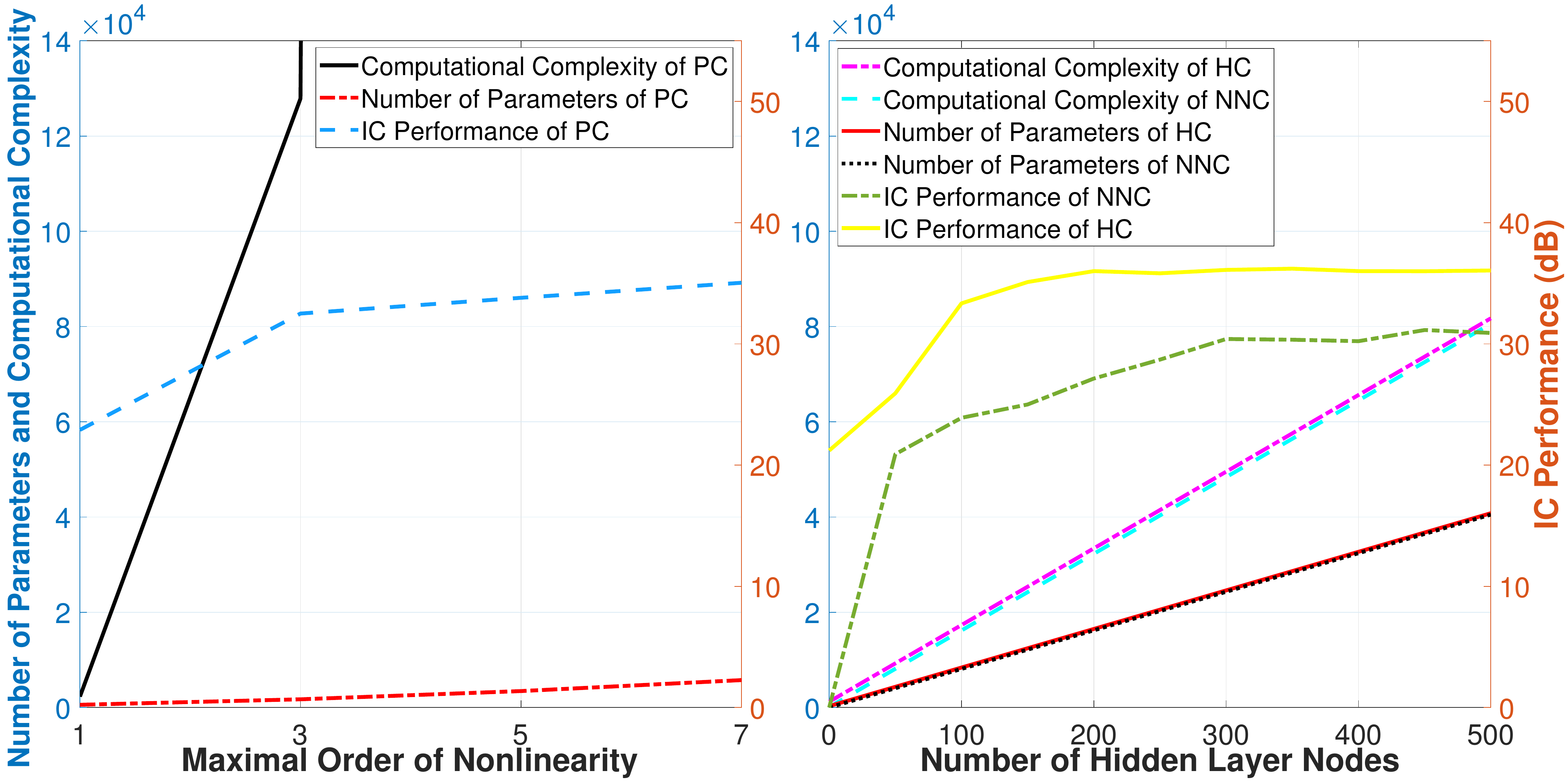}}
\vspace{-1pt}
\caption{The number of parameters, the computational complexity, the IC performance of the PC as a function of the order of nonlinearity (left), and those of the NNC and HC as a function of the number of hidden nodes (right).}
\vspace{-8pt}
\label{R1}
\end{figure}

In Fig. \ref{R1}, on the one hand, it is observed that the number of parameters of the PC increases linearly with the  order of nonlinearity $P$ assumed, while its computational complexity grows exponentially with $P$. Moreover, the number of parameters and the computational complexity of both the ML based cancellers increase linearly with the number of hidden nodes $N_h$. On the other hand, we characterize how the IC performance of different CLI cancellers changes with $P$ and $N_h$. For the PC, $P\!=\!3$ is used, since a larger $P$ leads to excessive computational complexity and slightly improved IC performance. For the NNC and HC, we set $N_\text{h}=300$ and $N_\text{h}=200$, respectively, to achieve high IC performance, small number of parameters to be estimated and low computational complexity. Therefore, the numbers of parameters of the CLI cancellers \textcolor{black}{are} $n_{\text {PC}}\!=\!1728$, $n_{\text {NNC}}\!=\!24310$ and $n_{\text {HC}}\!=\!16498$, respectively, while their computational \textcolor{black}{complexities} \textcolor{black}{are} $c_{\text {PC}}\!=\!\textcolor{black}{127864}$, $c_{\text {NNC}}\!=\!48380$ and $c_{\text {HC}}\!=\!\textcolor{black}{33424}$, respectively.

\begin{figure}[t]
\setlength{\abovecaptionskip}{-0.1cm}   
\setlength{\belowcaptionskip}{-5cm}  
\centerline{\includegraphics[width = .36\textwidth]{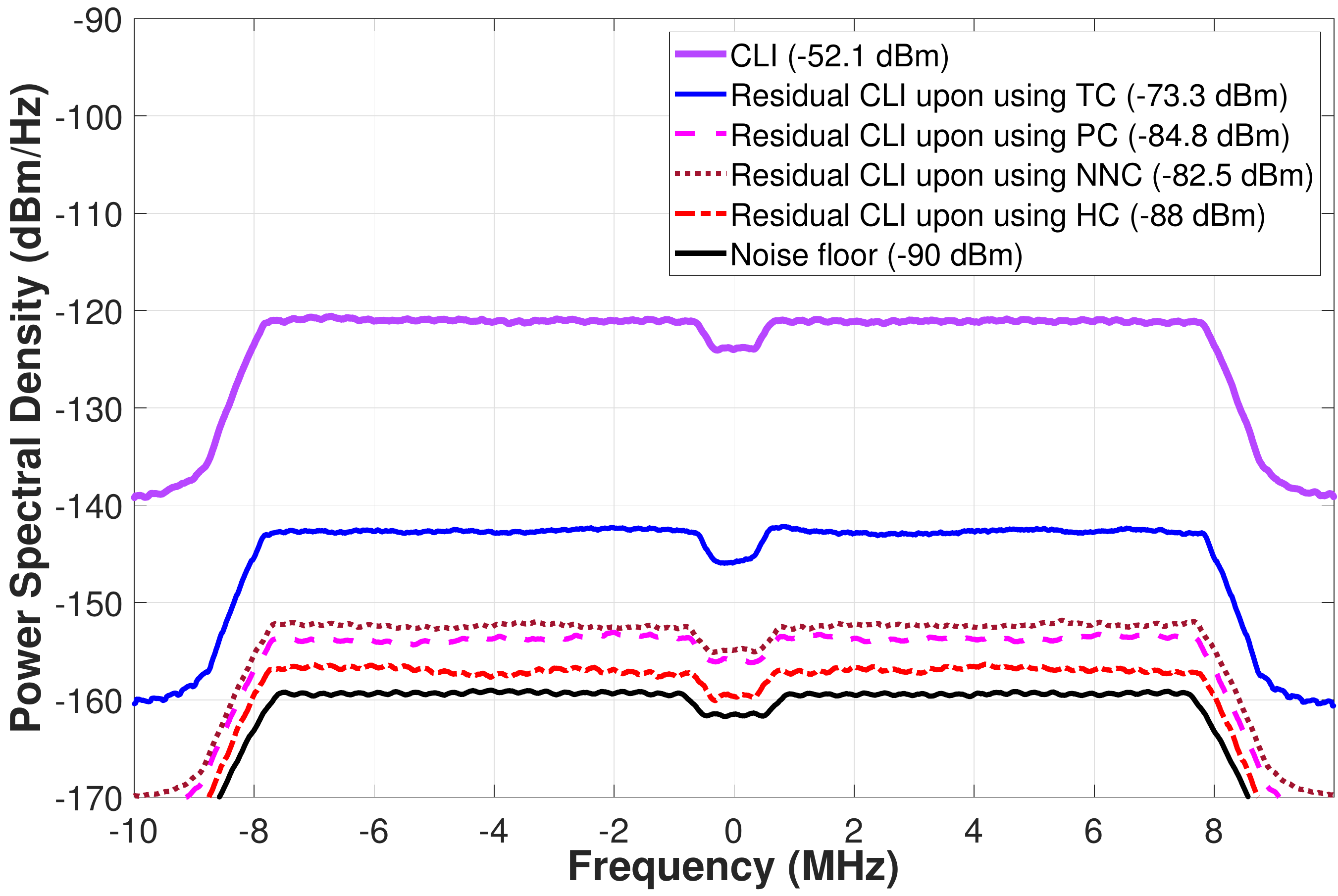}}
\vspace{-2pt}
\caption{Comparison of the IC performance in terms of the residual CLI power upon using different CLI cancellers, against the power of the received CLI and of the noise floor, which is the variance of ${n}^{n_0}_{0,k}$ in (1).}
\vspace{-17pt}
\label{R2}
\end{figure}

In Fig. \ref{R2} we compare the IC performance in terms of the residual CLI power upon using different CLI cancellers. The noise floor power is $-90$dBm, which is the lower limit of the residual CLI power. The power of CLI  received by the interfered BS$_0$ is $-52.1$dBm, which is the upper limit of the residual CLI power. A residual CLI of $-73.3$dBm remains upon using the TC, thus \textcolor{black}{the CLI cancellation performance of $C_{\mathrm{dB}}\!\!=\!\!21.2\text{dB}$} is achieved by the TC. \textcolor{black}{In} contrast, the CLI cancellation \textcolor{black}{performances} of the PC, the NNC and the HC \textcolor{black}{are} $C_{\mathrm{dB}}\!\!=\!\!32.7\text{dB}$, $C_{\mathrm{dB}}\!\!=\!\!30.4\text{dB}$ and $C_{\mathrm{dB}}\!\!=\!\!35.9\text{dB}$, respectively. 

\begin{figure}[t]
\vspace{-20pt}
\setlength{\abovecaptionskip}{-0.1cm}   
\setlength{\belowcaptionskip}{-5cm}  
\centerline{\includegraphics[width = .4\textwidth]{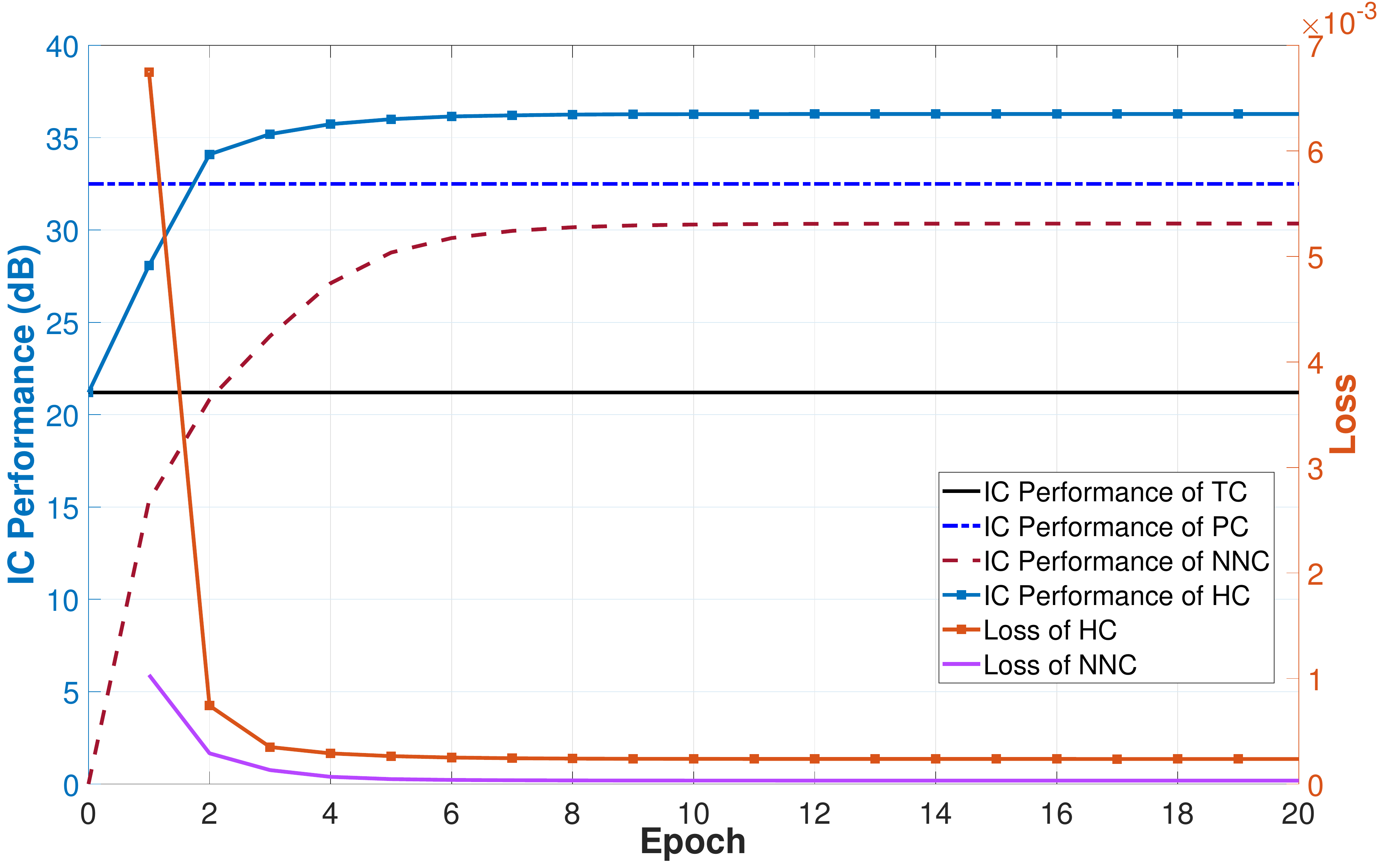}}
\vspace{-3pt}
\caption{The achievable IC performance ($C_{\mathrm{dB}}$) (left axis), and the loss of the NNC and the HC (right axis), as a function of the number of epochs. \textcolor{black}{The $\mathsf{loss}$ functions of NNC and HC are expressed as $\frac{1}{N}\sum_{n=1}^{N}\left \|{\mathbf{y}_{\text{NNC}}[n]}/{m_2^\text{NNC}}-\mathbf{\hat{y}}_{\text{NNC}}[n]'  \right \| _2^2 $ and $\frac{1}{N}\sum_{n=1}^{N}\left \|{\mathbf{y}_{\text{HC}}[n]}/{m_2^\text{HC}}-\mathbf{\hat{y}}_{\text{HC}}[n]'  \right \| _2^2 $, respectively.}}
\label{EpochRes}
\vspace{-17pt}
\end{figure}

In Fig. \ref{EpochRes}, we observe that the NNC and the HC converge and reach the peak of the IC performance after training a small number of epochs, which facilitates the deployment of the NNC and HC methods in realistic networks.
\vspace{-5pt}
\section{Conclusion}\label{Sc5}
We first propose a more realistic BS-to-BS channel model accounting for the effects of both the channel linearity and nonlinearity associated with the RF chain. Based on this model we propose three digital-domain CLI cancellers, namely the PC, NNC and HC. We demonstrate through analysis and numerical simulations that the three cancellers are able to achieve at least 43.4\% CLI cancellation performance improvement compared with the traditional CSI-based interference canceller. Among them, the PC has the least number of parameters. Compared with that of the PC, the computational \textcolor{black}{complexities} of both lightweight-ML based cancellers \textcolor{black}{are} reduced by more than \textcolor{black}{62.2\%}, and the HC, which deals with the linear and nonlinear CLI components differentially, has the highest CLI cancellation performance  ($C_{\mathrm{dB}}=35.9\text{dB}$) and the lowest computational complexity.

\ifCLASSOPTIONcaptionsoff
\newpage
\fi

\bibliographystyle{IEEEtran}
\bibliography{IEEEabrv,IEEEexample}

\end{document}